\documentclass[10pt, a4paper]{article}
\usepackage[latin1]{inputenc}
\usepackage[T1]{fontenc}
\usepackage{csquotes}
\usepackage[british,UKenglish,USenglish,english,american]{babel}
\usepackage{amssymb,amsmath,amsthm, mathrsfs}
\usepackage{dsfont}
\usepackage{hyperref}
\usepackage{tikz}
\usetikzlibrary{arrows}

\usepackage{algpseudocode}
\usepackage[linesnumbered,ruled,vlined]{algorithm2e}
\usepackage{graphicx}
\newtheorem{theorem}{Theorem}[section]

\newtheorem{exe}[theorem]{Example}
\newtheorem{defi}[theorem]{Definition}
\usepackage{amsmath}
\begin{document}

\title{\LARGE{Nonparametric adaptive active learning under local smoothness condition}}

\author{Boris Ndjia Njike, Xavier Siebert \\
\small Universit\'{e} de Mons,
\small Facult\'e polytechnique\\
 \small D\'{e}partement de Math\'{e}matique et recherche op\'erationnelle \\
 \small 
 \small e-mail: {\tt borisedgar.NDJIANJIKE@umons.ac.be}\\
%\and Merlin N. Chanter\\
%\small Tintagel University,
 %\small Roundtable Institute\\
 \small e-mail: {\tt xavier.siebert@umons.ac.be}
 % followed by \and and other authors with affiliations if needed
}

\date{}

\maketitle

\begin{abstract}
Active learning is typically used to label data, when the labeling process is expensive. Several active learning algorithms have been theoretically proved to perform better than their passive counterpart. However, these algorithms rely on some assumptions, which themselves contain some specific parameters. This paper adresses the problem of adaptive active learning in a nonparametric setting with minimal assumptions. We present a novel algorithm that is valid under more general assumptions than the previously known algorithms, and that can moreover adapt to the parameters used in these assumptions. This allows us to work with a larger class of distributions, thereby avoiding to exclude important densities like gaussians.
Our algorithm achieves a minimax rate of convergence, and therefore performs almost as well as the best known non-adaptive algorithms. 
%especially for those that do not implicitly simultaneous require that the density function of the marginal distribution of the instance space exists and is bounded away from zero on its support.
\end{abstract}

\section{Introduction}
\label{introduction}

The paradigm of passive learning consists in providing a classifier based on labelled data, identically independently distributed from a large pool of data. Due to a huge increase in the volume of the data available, we are sometimes constrained, from the point of view of the process of labeling data only,  to look beyond the standard passive learning. In this context, one of the most studied technique is active learning,  where the algorithm is presented with a large unlabelled pool of data and can iteratively request at a certain cost, the label $Y$ $\in$ $\lbrace 0,1\rbrace$ of an instance $X$ $\in$ $\mathbb{R}^d$ from the pool. We are constrained to use at most a budget of $n$ requests to a so-called oracle. The goal is to use this interaction to drastically reduce the number of labels needed to provide a classifier whose excess error is as small as possible.\\
Over the past decade, there has been a large body of work aiming at understanding theoretically the benefits and limits of active learning over passive learning \cite{dasgupta2006coarse, castro2006upper, dasgupta2011two}. One of the seminal works due to Castro and Nowak \cite{castro2006upper} analyzed various scenarios and provided one in which active learning outperforms passive learning. This situation corresponds to a common assumption called the Tsybakov noise assumption \cite{mammen1999smooth} that characterizes the noise near the boundary decision. Together with a smoothness assumption related to the boundary decision 
% %(and that ensures for example that two closest points, with respect to a specific metric ''tend'' to have the same label)
they provided an active learning strategy  that is better than the passive learning, in the sense that it uses fewer labels request to reach a low error. Also, in the parametric setting, Castro and Nowak \cite{castro2006upper} studied the effectiveness of  active learning for one-dimensional threshold classifier. Under the Tsybakov noise assumption, according to the knowledge of certain noise parameters, they provided an active learning algorithm, more effective than passive learning. However, one of the practical limitations of these active learning strategies is that the knowledge of the noise and smoothness parameters is required \cite{castro2006upper}.
% \textcolor{red}{in nonparametric setting 
%(on n'a pas encore parle de la distinction param / non param)}. %This sounds unrealistic in many cases in practice, so that it would be interesting to provide algorithms that adapt to these parameters.
This sounds unrealistic in many cases in practise, so that it would be interesting to provide algorithms that adapt to these parameters.\\

This paper is organized as follows. In Section~\ref{sec:state_of_the_art}, we provide a review of the main adaptive active learning algorithms, both in parametric and nonparametric settings.
In Section~\ref{sec:related_work} we describe some related works that inspired us and highlight the main contributions of our work. In Section~\ref{sec:defs} we provide the main definitions that will be used throughout this work. 
In Section~\ref{sec:assumptions} we explain the different assumptions and highlight their practical implications. In Section~\ref{sec:algo} we describe our new adaptive algorithm called AKALLS. Section~\ref{sec:minimax} provides upper and lower bounds on the excess risk of our algorithm. Section~\ref{sec:conclusion} is the conclusion of the paper.

\section{Existing work in adaptive active learning}
\label{sec:state_of_the_art}
In a parametric setting, Hanneke \cite{hanneke2011rates} opened the possibility for adaptation to certain key parameters like noise parameters by extending the work of Castro \textit{et al.} \cite{castro2008minimax} to a general class of hypotheses with finite complexity (VC-class, finite disagreement coefficient). Active learning strategies were designed on general classes of hypotheses and these active learning strategies were proved to adapt to the noise parameters \cite{hanneke2011rates}. In particular, one of these adaptive active learning algorithms achieves the same minimax rate as in the problem of learning a threshold classifier studied in \cite{castro2008minimax}. Also, Balcan and Hanneke \cite{balcan2012robust} introduced some theoretical aspects of a variant of standard active learning. Their algorithm allows to select an unlabelled subset of the pool and then to request a point that has a given label within this subset, if one exists. Under the Tsybakov noise assumption, some algorithms adaptive to the noise parameter were designed \cite{balcan2012robust}  based on a general class of hypotheses with finite complexity  (finite disagreement coefficient, VC-class or more generally finite Natarajan dimension). The rate of convergence achieved is as good as in non-adaptive setting, up to a logarithmic factor.\\

In nonparametric setting, Minsker \cite{minsker2012plug} assumed that the regression function $\eta(X)= \mathbb{E}(Y\vert X)$ belongs to the H\"older class $($with a fixed parameter $\alpha)$, and satisfies the Tsybakov noise assumption $($with a fixed parameter $\beta)$. By using a geometrical assumption called \textit{strong density assumption}, and under the condition $\alpha\beta\leq d$, he designed an adaptive active learning strategy that nearly achieves the minimax rate of convergence $\displaystyle n^{-\frac{\alpha(\beta+1)}{2\alpha+d-\alpha\beta}}$ better than in passive learning $\displaystyle n^{-\frac{\alpha(\beta+1)}{2\alpha+d}}$ where $n$ is the number of labels requested and $d$ the dimension of the instance space. However, Minsker's active learning strategy works with an additional assumption compared to the passive setting : the regression function relating the $L_2$ and $L_{\infty}$ approximation losses of certain piecewise constant or polynomial approximations of the regression function in the vicinity of the decision boundary. This algorithm is based on a model selection related to a dyadic partition of the cube $[0,1]^d$, where $d$ is the dimension of the data space, and he used a powerful oracle inequality that allows adaptation to the smoothness parameter $\alpha$. Remarkably, this algorithm adapts naturally to the noise parameter $\beta$.\\
 Locatelli \textit{et al.} \cite{locatelli2017adaptivity} also consider a dyadic partition along with the H\"older smoothness and the Tsybakov noise assumption on the regression function. By using the strong density assumption,  and under the condition $\alpha\beta\leq d$, they provided an active learning strategy that adapts both to the smoothness and naturally to the noise parameters and that achieves the same minimax rate as obtained in \cite{minsker2012plug}. They assumed that the smoothness parameter $\alpha$ belongs to a range of values $I$ and considered a finite increasing sequence $(\alpha_i)\subset I$. Their adaptive algorithm with respect to the smoothness parameter is based on a non-adaptive algorithm that iteratively takes as input a smoothness parameter $\alpha_i$ $(i=1,2,...)$ and outputs a labeled set $\mathcal{S}_i$. Because the H\"older class is a nested class, the label of a point does not change between two consecutive iterations and then $\mathcal{S}_i\subset \mathcal{S}_{i+1}$. Finally, Locatelli \textit{et al.} proved that it is possible to control the error rate beyond the maximum $\alpha_i$ such that $\alpha_i\leq \alpha$ and obtained the optimal rate of convergence up to a logarithmic factor.\\

In the context of nonparametric active learning under a smoothness assumption, the problem of designing adaptive algorithms that achieves optimal rates under minimal assumption is still evolving. In this paper, we aim  at designing an adaptive active learning strategy that achieves an optimal rate, but under a more general smoothness assumption than that used previously  \cite{minsker2012plug, locatelli2017adaptivity}.

\section{Related work and new contributions}
\label{sec:related_work}
%\subsection{Related work in passive learning}
Chaudhuri and Dasgupta \cite{chaudhuri2014rates} studied the problem of passive learning (more specifically, K-nn classification) under minimal assumptions. Their motivation was to design a smoothness assumption related to the underlying marginal density $P_X$ which therefore allows to overcome some disadvantages of the H\"older smoothness assumption. 
%From the point of view of applications, this new smoothness assumption is suitable for the K-nn classification that is amongst the simplest classification rules. 
Under this new smoothness assumption they provided a K-nn classifier, and designed a region of confidence that can reliably be classified, and outside of which the error rate is controlled by the Tsybakov noise assumption. This allows to achieve a rate of convergence as good as under H\"older smoothness in passive learning. 

%\subsection{Our previous contribution in active learning}
This work was previously extended to the context of active learning \cite{ndjia2020k}. Under the new smoothness assumption, and the Tsybakov noise assumption, an active learning algorithm was designed, which achieves the same rate of convergence as was obtained under the H\"older smoothness assumption in \cite{locatelli2017adaptivity, minsker2012plug}. This algorithm is based on a pool of unlabeled examples $\mathcal{K}$, and consists in providing a labeled subset $\hat{S} \subset \mathcal{K}$ called the \textit{active set} and finally consider the 1-NN classifier on $\hat{S}$. Instead of asking directly the label of an example in $\hat{\mathcal{S}}$, it infers it by asking the labels of its neighbors and then it obtains the correct label for a point relatively far from the Bayes decision boundary. Finally, \cite{ndjia2020k} proved that for each example in the interior of the support of the underlying marginal distribution, relatively far away from the Bayes decision boundary $\lbrace x,\;\eta(x)=\frac12\rbrace$, its label coincides with both the true label and the inferred label of its nearest neighbor in $\mathcal{S}$. But, for a practical point of view, their algorithm may be sometimes not applicable due to the fact that it requires the knowledge both to the smoothness and noise parameters. 
%In this paper, we provide a new adaptive algorithm, which competes with the best algorithms that assume the knowledge of the noise and smoothness parameters.
\subsection{Contributions of this paper}
In this work, we establish two main results.
First, we provide an active learning algorithm that adapts both to the smoothness and noise parameters, and prove theoretically that it achieves the same rate of convergence as that of non-adaptive algorithms which require the knowledge of smoothness and  noise parameters. It is important to underline that our smoothness assumption is more general than was done in the previous works, particularly the H\"older  smoothness assumption.
Second, we also extend the work of \cite{ndjia2020k} by providing a lower bound that matches (up to a logarithmic factor) the upper bound established in this paper. 

\section{Setting and definitions}
\label{sec:defs}
\subsection{Active learning setting}
\label{sec:ALsetting}
Let $\mathcal{X}\subset \mathbb{R}^d$ the data space, called instance space and $\mathcal{Y}=\lbrace 0,1\rbrace$ the label space. Let $w$ $\in$ $\mathbb{N}^*$ and an i.i.d sample $\mathcal{K}\subset \mathcal{X}\times \mathcal{Y}$:
$$\mathcal{K}=\lbrace (X_1,Y_1),\ldots, (X_w,Y_w)\rbrace$$ drawn according to a probability $P$ over $\mathcal{X}\times \mathcal{Y}$ and $$\mathcal{K}_x=\lbrace X_1,\ldots, X_w\rbrace$$ its corresponding sequence of unlabeled points
. The probability $P$ can be decomposed as a couple $(P_X, \eta)$ where $P_X$ is the marginal probability on $\mathcal{X}$ and $\eta$ the regression function defined by $\eta(x)=P(Y=1\vert X=x)$ for all $x$ in the support of $P_X$. We define a classifier as a measurable function $f:\mathcal{X}\rightarrow \mathcal{Y}$. The standard (passive) learning based on the sample $\mathcal{K}$ consists in designing an algorithm that provides a classifier $\hat f_w$. The performance of $\hat f_w$ is measured by $R(\hat f_w)$ where the function $R$ is called classification error, is defined by $R(f)=P(f(X)\neq Y)$ over all measurable functions $f:\mathcal{X}\rightarrow \mathcal{Y}$. 
It is known \cite{lugosi2002pattern} that the  Bayes classifier, defined by $f^*(x)=\mathds{1}_{\eta(x)\geq 1/2}$ minimizes the classification error. Then for a classifier $f$, the quantity $R(f)-R(f^*)$ is called the excess risk of $f$. \\In active learning, we do not directly have access to the label of $X$ $\in$ $\mathcal{K}_x$ and requesting its label is considered costly. At the beginning, the label budget $n$ is thus fixed. The challenge consists in designing a strategy by requesting at most $n$ labels while achieving a performance competitive with that of passive learning, where the label budget would correspond to $n=w$. At any time, we choose to request the label of a point $X$ $\in$ $\mathcal{K}_x$ according to the previous observations. The point $X$ is chosen to be most \enquote{informative}, which amounts to belonging to a region where classification is difficult and requires more labeled data to be collected.   

\subsection{Definitions}
In this section, we present some definitions of the important concepts we use throughout this paper. First let us recall $\mathcal{X}\subset \mathbb{R}^d$ the instance space, and $\rho$ the Euclidean metric on $\mathcal{X}$. For $x$ $\in$ $\mathcal{X}$, and $r>0$,  we define $\bar B(x,r)=\lbrace z\in \mathbb{R}^d,\; \rho(z,x)\leq r\rbrace$ and $B(x,r)=\lbrace z\in \mathbb{R}^d,\; \rho(z,x)< r\rbrace$.
\begin{defi}{$(\alpha,L)$-H\"older smoothness}~\\
\label{def:smooth1}
Let $\eta: \mathcal{X}\rightarrow [0,1]$ the regression function (defined in Section~\ref{sec:ALsetting}). We say that $\eta$ is $(\alpha,L)$-{\bf{H\"{o}lder continuous}} $(0<\alpha\leq 1, \text{and}\; L>1)$ if $\forall$ $x,x'\in \mathcal{X}$,   
\begin{equation}
\label{def:Holder}
\vert \eta(x)-\eta(x')\vert\leq L\rho(x,x')^{\alpha}. \tag{H1} 
\end{equation}
\end{defi}

\begin{defi}{$(\alpha,L)$-smoothness}~\\
\label{def:smooth2}
Let $0<\alpha\leq 1$ and $L>1$. The regression function is \textbf{$(\alpha,L)$-smooth} if for all $x,x'$ $\in$ supp$(P_X)$ we have: 
\begin{equation}
\label{def:smooth}
\begin{split}
&\vert \eta(x)-\eta(x')\vert \leq L.P_X(B(x,\rho(x,x')))^{\alpha/d},   
\end{split} 
\tag{H2}
\end{equation}
where $d$ is the dimension of the instance space.
%where $$\eta(B(x,r))=\frac{1}{P_X(B(x,r))}\int_{B(x,r)} \eta(x')dP_X(x').$$
\end{defi} 

\begin{defi}{Margin noise}\label{Tsy}~\\
We say that $P$ satisfies \textbf{margin noise} or \textbf{Tsybakov's noise} assumption with parameter $\beta\geq 0$ if for all $0<\epsilon\leq 1$ 

\begin{equation}
\label{def:TsybakovMarginNoise}
P_X(x\in\mathcal{X},\;\vert\eta(x)-1/2\vert<\epsilon)<C\epsilon^{\beta}, \tag{H4} 
\end{equation} for  $C:=C(\beta)\in [1,+\infty[$.

\end{defi}
\begin{defi}{Strong density}\label{def:strong}~\\
Let $P$ the distribution probability defined over $\mathcal{X}\times \mathcal{Y}$ and $P_X$ the marginal distribution of $P$ over $\mathcal{X}$. We say that $P$ satisfies the \textbf{strong density} assumption if there exists some constants $r_0>0$, $c_0>0$, $p_{min}>0$ such that for all $x$ $\in$ $\text{supp}(P_X)$: 
\begin{equation}
\begin{split}
&\lambda(B(x,r)\cap \text{supp}(P_X))\geq c_0 \lambda(B(x,r)),\;\forall r\leq r_0\\
&\text{and}\; p_X(x)>p_{min},
\end{split}
\tag{H3}
\label{def:strongDensity}
\end{equation}
where $p_X$ is the density function of the marginal distribution $P_X$ and $\lambda$ is the Lebesgue measure.
\end{defi} 

\section{Assumptions}
\label{sec:assumptions}
In this work, we use two main assumptions described in details in this section.
\subsection{First assumption}
\hypertarget{smooth}{\textbf{\underline{Assumption 1:}}} We suppose that the regression function  satisfies \eqref{def:smooth}. \\~\\
This assumption was introduced by Chaudhuri and Dasgupta \cite{chaudhuri2014rates}, who pointed out some disadvantages of the assumption \eqref{def:Holder}. Their motivation was to define a smoothness assumption that measures the change of the regression function with respect to the marginal distribution $P_X$, instead of the H\"older smoothness assumption \eqref{def:Holder} that measures the change of the regression function with respect to the instance $x$. They also proved that \eqref{def:smooth} generalizes \eqref{def:Holder} along with \eqref{def:strongDensity} as stated in the following theorem.

\begin{theorem}[Chaudhuri and Dasgupta]\cite{chaudhuri2014rates}\label{theo1}~\\ 
Suppose that $\mathcal{X}\subset\mathbb{R}^d$, that the regression function $\eta$ is $(\alpha_h,L_h)$-H\"older smooth, and that $P_X$ satisfies \eqref{def:strongDensity}. Then there is a constant $L>1$ such that for any $x,z$ $\in$ supp($P_X$), we have: 
$$\vert \eta(x)-\eta(z)\vert \leq L.P_X(B(x,\rho(x,z)))^{\alpha_h/d}.$$
\end{theorem} 

This theorem states that a regression function which satisfies \eqref{def:Holder} and \eqref{def:strongDensity} also satisfies \eqref{def:smooth}.

To illustrate the importance of this assumption, we provide an example of a regression function that does not satisfy simultaneously \eqref{def:Holder} and \eqref{def:strongDensity}, but satisfies \eqref{def:smooth}. 

\begin{exe}[Distribution that satisfies \eqref{def:smooth}]~\\
Let $P=(\eta,P_X)$ the distribution defined as follows: 
\begin{itemize}
\item The marginal distribution $P_X$ is such that $X\sim\mathcal{N}(0,1)$ the univariate normal distribution.
\item For $\alpha\leq 1$ the regression function is defined by:  

\begin{align*}
\eta:\;&\mathbb{R}\longrightarrow [0,1]\\
     & x\longmapsto 
     \left\{
    \begin{array}{ll}
        1-\tfrac{2^{\alpha+1}}{3}\vert x-\frac 1 2\vert^{\alpha}\;\;\text{if}\;\; x\in[0, 1]\\~\\
        \frac 13 \;\;\text{elsewhere}.
    \end{array}
\right.
\end{align*}
This regression function is represented simultaneously with the density function of the univariate normal distribution on Figure~\ref{fig:ex}.
\end{itemize}
%\shorthandoff{:}

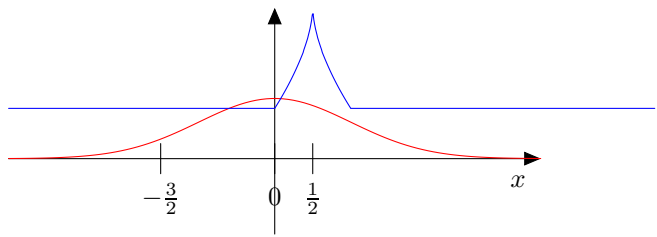
\begin{figure}[ht!]
\begin{center}
\begin{tikzpicture}[line cap=round,line join=round,>=triangle 45,x=1cm,y=1cm]
\begin{scope}[yscale=2]
\draw [->] (-3.5,0) -- (3.5,0);
\foreach \x/\xtext in {-1.5/-\frac 32,0, 0.5/\frac 12} {\draw (\x,0.1cm) -- (\x,-0.1cm) node[below] {$\xtext\strut$};}
\draw [->] (0,-0.5) -- (0,1);
\draw[domain=-3.5:3.5,samples=100,color=red] plot ({\x},{1/(sqrt(2*pi))*exp(-(\x*\x)/2)});
\draw[domain=0:1,samples=100,color=blue] plot ({\x},{1-2^(0.6+1)/3*abs(\x-1/2)^(0.6)});
\draw[domain=-3.5:0,samples=100,color=blue] plot ({\x},{1/3});
\draw[domain=1:5,samples=100,color=blue] plot ({\x},{1/3});
\clip(-2,-0.3) rectangle (4,1.3);
\draw (3.2,-0.15) node {$x$};
%\draw (-0.4,1) node {$\eta(x)$};
\end{scope}
\end{tikzpicture}
\caption{Example of regression function $\eta(x)$ (blue) that satisfies \eqref{def:smooth} along with the marginal distribution $P_X$ (red).}
\label{fig:ex}
\end{center}
\end{figure}

%\shorthandon{:}
The probability $P$ does not satisfy \eqref{def:strongDensity} because the marginal density is not bounded below and it can easily be shown that it satisfies \eqref{def:smooth} with parameters $(\alpha, \frac 2 3\exp(-\frac 12\alpha))$ or more formally $(\alpha, 1)$ because the constant $L$ in \eqref{def:smooth} is greater than 1.
\end{exe}

\subsection{Second assumption}
\hypertarget{noise}{\textbf{\underline{Assumption 2:}}} (Tsybakov noise assumption)\\~\\
We suppose that $P$ satisfies the Tsybakov noise assumption with parameters $(\beta, C)$ such that $\beta>0$, $C>1$.\\
This assumption was introduced in \cite{mammen1999smooth} and characterizes the behavior of the regression function near the decision boundary using a parameter $\beta$. For a large value of $\beta$, we can observe  a "jump" of the regression function on the decision boundary, and a small value of $\beta$ covers the interesting case where the regression function crosses the decision boundary.

\section{The AKALLS algorithm}
\label{sec:algo}
In Section~\ref{sec:overall} we provide a general description of the AKALLS algorithm. Then in Section~\ref{sec:notations} we introduce some notations that will be used through the remainder of this paper. The pseudo-code of AKALLS algorithm is provided in Section~\ref{sec:AKALLS}, and the main subroutines are explained in Section~\ref{sec:subroutines}.

\subsection{Overall Description}
\label{sec:overall}
Our active learning algorithm adapts to the smoothness and noise parameters ($\alpha$ and $\beta$, respectively), at least in a reasonable range of these parameters.

The algorithm takes as input a pool of unlabelled data $\mathcal{K}$, a label budget $n$, the constant parameters $L$, $C$ respectively used in  \hyperlink{smooth}{Assumption 1} and \hyperlink{noise}{Assumption 2}, a confidence parameter $\delta$ $\in$ $(0,1)$, an accuracy parameter $\epsilon$ $\in$ $(0,\frac 12)$. For handling the adaptivity to the parameters $L$, $C$, we suppose they are both bounded by a logarithmic factor in $\frac{1}{\epsilon}$. 

We design a decreasing sequence of smoothness parameters  $(\alpha_i)$ such that at each step $i$, we execute a non-adaptive algorithm similar to that introduced in \cite{ndjia2020k}. 

Each step produces a set $\widehat{\mathcal{S}}_i$ of informative points. The sequence $(\widehat{\mathcal{S}}_i)$ is increasing, and at the end of step $i$, the points added in $\widehat{\mathcal{S}}_i$ potentially improve the classification compared to the previous step. At the end of our algorithm, we obtain an aggregate set $\mathcal{S}$ on which we apply a 1-NN classifier.

%\newpage
\subsection{Notations}
\label{sec:notations}
For $X_s$ $\in$ $\mathcal{K}=\lbrace X_1,\ldots, X_w\rbrace$, we denote by $X^{(k)}_s$ its $k$-th nearest neighbor in $\mathcal{K}$, and $Y^{(k)}_s$ the corresponding label.

For an integer $k\geq 1$, let 
\begin{equation}
\label{eq:eta_Xs}
\widehat{\eta}_k(X_s)=\frac 1k\sum_{i=1}^{k} Y_{s}^{(i)},\quad \bar\eta_k(X_s)=\frac 1k\sum_{i=1}^{k} \eta(X_{s}^{(i)}).
\end{equation}
For a set $\mathcal{S}\subset \mathcal{X}\times \mathcal{Y}$, we denote by $\mathcal{S}_x$ the set $$\mathcal{S}_x=\lbrace X\in \mathcal{X},\;\;(X,Y)\,\in\, \mathcal{S}\rbrace.$$

Let us fix $\epsilon,\;\delta,\;\Delta \in (0,1)$. The following quantities \eqref{eq:k_s}, \eqref{eq:pi_ks}, \eqref{eq:margin1}, \eqref{eq:sigma} are derived from the detailed convergence proofs.
\begin{equation}
k(\delta,\Delta)=\frac{c}{\Delta^2}\left[\log(\frac{1}{\delta})+\log\log(\frac{1}{\delta})+\log\log\left(\frac{512\sqrt{e}}{\Delta}\right)\right]
\label{eq:k_s}
\end{equation}
where $c\geq 7.10^6$.\\

\begin{equation}
b_{\delta,k}=\sqrt{\frac 2k\left(\log\left(\frac{1}{\delta}\right)+ \log\log\left(\frac{1}{\delta}\right)+ \log\log(ek)\right)}.
\label{eq:pi_ks}
\end{equation}

\begin{equation}
\Delta_0=\max(\frac{\epsilon}{2}, \left(\frac{\epsilon}{2C}\right)^{\frac{1}{\beta+1}})
\label{eq:margin1}
\end{equation}
where $(\beta,C)$ are the parameters introduced in \hyperlink{noise}{Assumption 2}.\\

\begin{equation}
\phi_n=\sqrt{\frac {1}{n}\left(\log\left(\frac{1}{\delta}\right)+ \log\log\left(\frac{1}{\delta}\right)\right)}.
\label{eq:sigma}
\end{equation}

%\textcolor{red}{ on avait dit que $\widehat\Delta=\tilde{O}(n^{-\frac{\alpha}{2\alpha+d-\alpha\beta}})$ -- expliquer !!!}

\subsection{AKALLS Algorithm}
\label{sec:AKALLS}
%\newpage
\begin{algorithm}[!htbp]
 \caption{Adaptive Active Learning under Local Smoothness (AKALLS)}
 \label{algo:AKALLS}
\KwIn{A pool $\mathcal{K}_x=\lbrace X_1,\ldots,X_w\rbrace$,  label budget $n$, $L$, $C$, confidence parameter $\delta$, accuracy parameter $\epsilon$.} 
\KwOut{1-nn classifier $\widehat{f}_{n,w}$}
\textbf{Initialization}\\
$\bar n=\frac{n}{\log_2(\frac{1}{\epsilon})}$\\
$I_0=\emptyset$\\
Current active set \; $\widehat{\mathcal{S}}_0=\emptyset$,\\ 
Current "noisy" points\; $\widehat{\mathcal{S}}_{nois}=\emptyset$\\
%$I=\emptyset$ \Comment{set of $"$informative points$"$; used for providing the label complexity}\\ 
$i=1$

\Repeat{$i>\log_2(\frac{1}{\epsilon})$}{

             $s=1$ \Comment{index of point currently examined}\\
             $t=\bar n$ \Comment{current label budget}\\
             $\alpha_i=2^{1-i}$\\
             $\widehat{\mathcal{C}}_{i}=\emptyset$ \Comment{current informative set at the i-th step}\\
             $I=\emptyset$\\
 
      \Repeat{$t<0$ and $s> w$}{ 
            \If{$X_s$ $\in$ $(\widehat{\mathcal{S}}_{i-1})_x\cup (\widehat{\mathcal{S}}_{nois})_x$}{
              $s=s+1$}
            \Else{
                 $T$=\texttt{Reliable}($X_s$, $\delta_s$, $\alpha_i$, $L$, $I\cup I_{i-1}$)}
               \If{T=True}{
                  $s=s+1$}
               \Else{
                   Let $\delta_s=\frac{\delta}{32s^2\log_2(\frac{1}{\epsilon})}$\\
                   $[\widehat{Y},Q_s]$=\texttt{confidentAdapt}($X_s$, $\epsilon$, $t$, $\delta_s$)\\
                   $\displaystyle \widehat{LB}_s=\left|\frac{1}{\vert Q_{s}\vert}\sum_{(X,Y)\in Q_{s}} Y-\frac 12\right|- b_{\delta_s,\vert Q_{s}\vert}$ \Comment{Lower bound guarantee on $\vert\eta(X_s)-\frac 12\vert $}\\
                   
                   $t=t-\vert Q_s\vert$}
                  \If{$\widehat{LB}_s\geq 0.1b_{\delta_s,\vert Q_s\vert}$}{
                    $\widehat{\mathcal{C}}_i=\widehat{\mathcal{C}}_{i}\cup \lbrace(X_s,\widehat{Y})\rbrace$\\
                    $I=I\cup \lbrace (X_s, \widehat{LB}_s)  \rbrace$}
                   \Else{
                     $\widehat S_{nois}=\widehat S_{nois}\cup \lbrace(X_s,\widehat{Y})\rbrace$}}

      $\widehat{\mathcal{S}}_i=\widehat{\mathcal{C}}_{i}\cup \widehat{\mathcal{S}}_{i-1}$\\
      $I_i=I\cup I_{i-1}$\\ 
      $i=i+1$}

$\mathcal{S}=\mathcal{S}_{\log_2(\frac{1}{\epsilon})}$\\
$\widehat{f}_{n,w}\leftarrow$ \texttt{Learn} $(\widehat{\mathcal{S}})$ 
\end{algorithm}

\newpage
\subsection{Main subroutines}
\label{sec:subroutines}
The AKALLS algorithm uses two main subroutines called \texttt{Reliable} and \texttt{ConfidentAdapt}.

The  \texttt{Reliable} subroutine is a boolean test that checks if the label of the current point $X_s$ can be inferred with high confidence using the information collected on the previous points examined by the subroutine \texttt{ConfidentAdapt}. If the \texttt{Reliable} subroutine returns True at point $X_s$,  the latter is not  considered to be informative, and therefore is not considered further by the subroutine \texttt{ConfidentAdapt}.
Conversely, if the \texttt{Reliable} subroutine returns False at point $X_s$,  the \texttt{ConfidentAdapt} subroutine is used to determine  its label with a given level of confidence. The \texttt{ConfidentAdapt} subroutine infers  the label of $X_s$ by using the labels of its nearest neighbors, with respect to a sequence of noise parameters $(\beta_i)$ in an adaptive way.

\subsubsection{\texttt{Reliable} subroutine}
The \texttt{Reliable} subroutine takes as inputs an instance point $X$, a confidence parameter $\delta$, the smoothness parameters $\alpha$, $L$, a set $I\subset \mathcal{X}\times\mathbb{R}$. For $(X',c)$ $\in$ $I$, $X'$ represents a point whose label we have already inferred with a guarantee $c$. Then the \texttt{Reliable} subroutine allows us to know if we can guess with high probability the label of point $X$, by using the set $I$.  The \texttt{Reliable} subroutine uses the marginal distribution $P_X$ which is supposed to be known by the learner. This is not a limitation, since we can assume that our pool of data is large enough such that $P_X$ can be estimated to any desired accuracy as was done in \cite{ndjia2020k}.

\begin{algorithm}[h!]
\caption{{\texttt{Reliable} subroutine}}
\label{algo:reliable}
\KwIn{an instance $X$, a confidence parameter $\delta$, smoothness parameters $\alpha$,  $L$, a set $I\subset \mathcal{X}\times\mathbb{R}$}
 KwOut{A boolean value $T$}
 
 \For{$ (X',c) \in I$}{
  
     \If{$\exists$ $(X',c)$  $\in$ $I$ such that $P_X(B(X,\rho(X,X'))\leq \left(\frac{c}{64L}\right)^{d/\alpha} $}{
        $T=True$}
     \Else{
         $T=False$}}
              
\end{algorithm}

\subsubsection{\texttt{ConfidentAdapt subroutine}} \texttt{ConfidentAdapt} takes as input an instance $X$, an accuracy parameter $\epsilon$, a budget parameter $t\geq 1$ and a confidence parameter $\delta$. \texttt{ConfidentAdapt} infers the label of an instance $X \in$ $\mathcal{K}_x$ by requesting the label of its neighbors in the pool $\mathcal{K}_x$. The output $\widehat{Y}$ corresponds to the majority vote of requested labels. \texttt{ConfidentAdapt} operates in adaptive way, so that we do not have to know beforehand the smoothness and the noise parameters.  Indeed, we introduce in the subroutine several noise levels $\beta_i$ and we expect that if the noise parameter $\beta\geq \beta_i$, \texttt{ConfidentAdapt} uses at most $k(\delta,\Delta_i)$ label requests.

\texttt{ConfidentAdapt} is designed such that the inferred label  produced at point $X_s$ with a given value of $\alpha_i$ does not change subsequently (for $\alpha_j$, $j> i$). Consequently, at iteration $i$ (relatively to the smoothness parameter $\alpha_i$), any point that has already been examined previously by \texttt{ConfidentAdapt} can no longer introduced into  the \texttt{Reliable} and \texttt{ConfidentAdapt} subroutines in the future iterations $(\alpha_j$, $j> i)$. 

\begin{algorithm}[h!]
\caption{{\texttt{confidentAdapt} subroutine}}
\label{algo:ConfidentAdapt}
\KwIn{an instance $X$, accuracy parameter $\epsilon$, budget parameter $t\geq 1$, confidence parameter $\delta$.}
\KwOut{$(\widehat{Y}, Q)$} 
\textbf{Initialization:}\\ 
$Q=\emptyset$\\ 
$k=1$     

\For{$i=1$: $\log^3(\frac{1}{\epsilon})$}{
    $\beta_i=\frac{i}{\log^2(\frac{1}{\epsilon})}$\\
    $\Delta_i=\max(\frac{\epsilon}{2}, \left(\frac{\epsilon}{2C}\right)^{\frac{1}{\beta_i+1}})$}
   
\For{$i=1$ to $\log^3(\frac{1}{\epsilon})$}{
    
    \Repeat{$k>\min(k(\delta,\Delta_i),t)$}{
           Request the label $Y^{(k)}$ of $X^{(k)}$\\
           $Q=Q\cup \lbrace (X^{(k)},Y^{(k)})\rbrace$\\ 
           
           \If{$\displaystyle\left\vert\frac{1}{k}\sum_{j=1}^{k}Y^{(j)}-\frac 12\right\vert>2b_{\delta,k}$}{
              exit\quad \Comment{cut-off condition}}
           
           $k=k+1$}}

$\displaystyle\widehat{\eta}\leftarrow\frac{1}{\vert Q \vert}\sum_{(X,Y)\in Q} Y$\\
$\widehat{Y}=\mathds{1}_{\widehat{\eta}\geq 1/2}$  
\end{algorithm}

\section{Bounds on AKALLS's Excess risk}
\label{sec:minimax}
In this section we provide the upper and lower bounds on the excess risk of our algorithm. We state these bounds in a more practical form by using label complexity.  

\subsection{Upper Bound}
In this Section we show that the rate of convergence achieved by \texttt{AKALLS} is nearly the same (up to a logarithmic factor) as that achieved by non-adaptive algorithms. It is important to note that this rate of convergence covers only the case $\alpha\beta\leq 1$, especially when the regression function crosses the boundary decision in the interior of the support of $P_X$.  Let us write $\mathcal{P}(\alpha,\beta):=$ the set of distribution of probabilities that satisfy \hyperlink{smooth}{Assumption 1} and \hyperlink{noise}{Assumption 2}, where the parameters $\alpha$ and $\beta$ respectively come from \eqref{def:smooth} and \eqref{def:TsybakovMarginNoise}.\\
 The following Theorem states the upper bound on the excess risk of the classifier provided by the \texttt{AKALLS} algorithm. 
\begin{theorem}~\\
\label{theo1}
Let $\epsilon$, $\delta$ $\in$ $(0,\frac 12)$, $n$ $\in$ $\mathbb{N}$, $d$ the dimension of the instance space. Let $\alpha$ $\in$ $(2\epsilon, 1)$, and $\beta$ $\in$ $[\frac{1}{\log^2(\frac{1}{\epsilon})},\;\log(\frac{1}{\epsilon})]$. Let $\mathcal{K}=\lbrace X_1,\ldots, X_w\rbrace$ a pool of data. There exists an active learning algorithm based on $\mathcal{K}$,  that is independent of $\alpha$ and $\beta$, which provides a classifier $f_{n,w}$ by using at most $n$ label requests such that:\\
 \textbf{If} $\alpha\beta\leq d$, and the number of labels request satisfies: 

\begin{equation}
\label{eq:label-complexity}
n\geq \tilde{O}\left( \left(\frac {1}{\epsilon}\right)^{\frac{2\alpha+d-\alpha\beta}{\alpha(\beta+1)}}\right),
\end{equation}

and $w$ satisfies 
\begin{equation}
\label{guarantee-on-pool-0}
w\geq \tilde{O}\left( \left(\frac{1}{\epsilon}\right)^{\frac{2\alpha+d}{\alpha(\beta+1)}}\right)
\end{equation}

\textbf{then} with probability at least $1-\delta$ we have: 
\begin{equation}
\label{eq:error}
\sup_{P\in\mathcal{P}(\alpha,\beta)}\,\left[ R(\widehat{f}_{n,w})-R(f^*)\right]\leq \epsilon.
\end{equation}
\end{theorem}
We can equivalently express Theorem~\ref{theo1} only as a function of number of label requests $n$. Specifically, there are  values $n$, and $w$ sufficiently large such that:

\begin{equation}
\label{eq:error1}
\sup_{P\in\mathcal{P}(\alpha,\beta)}\,\left[ R(\widehat{f}_{n,w})-R(f^*)\right]\leq \tilde{O}\left(n^{-\frac{\alpha(\beta+1)}{2\alpha+d-\alpha\beta}}\right).
\end{equation}  

\subsection{Minimax Lower Bounds}

In this Section we state that for a given probability $P$ $\in$ $\mathcal{P}(\alpha,\beta)$, no active learner can provide a classifier whose expected excess risk (with respect to the sample)  decreases to 0 faster than $\tilde{O}\left(n^{-\frac{\alpha(\beta+1)}{2\alpha+d-\alpha\beta}}\right)$. Combined with \eqref{eq:error1}, this therefore provides a minimax rate on the form: 
$$\tilde{O}\left(n^{-\frac{\alpha(\beta+1)}{2\alpha+d-\alpha\beta}}\right).$$
The following theorem is inspired by the minimax bounds of  \cite{audibert2007fast, minsker2012plug,locatelli2017adaptivity}. 

\begin{theorem}~\\
Let $\alpha$, $\beta$ the smoothness and noise parameters respectively introduced in \ref{def:smooth}, and \ref{def:TsybakovMarginNoise} and $d$ the dimension of the instance space. Let us assume that $\alpha\beta\leq  d$ and for any $P$ $\in$ $\mathcal{P}(\alpha,\beta)$, $supp(P_X)\subset[0,1]^d$. Then there exists a constant $\gamma>0$  such that for all $n$ large enough and for any active classifier $\widehat{f}_n$ we have: 

\begin{equation}
\label{eq:error2}
\sup_{P\in\mathcal{P}(\alpha,\beta)}\,\left[ R(\widehat{f}_{n,w})-R(f^*)\right]\geq \gamma n^{-\frac{\alpha(\beta+1)}{2\alpha+d-\alpha\beta}}.
\end{equation}
\end{theorem}
%\section{Discussion} 

\section{Conclusion} 
\label{sec:conclusion}

In this paper, we described an active learning algorithm with minimal regularity assumptions, that adapts to the parameters  used in these assumptions.
This algorithm achieves a better rate of convergence than its passive counterpart.  Additionally, we provided a lower bound on the excess risk, and therefore obtained a minimax rate of convergence. Interesting future directions include an extension to multi-class instead of binary classification. Also, due to the computational issues in high-dimensional feature spaces, we could assume that the data is constrained to a lower-dimensional manifold, a setting in which the nearest neighbors method of our algorithm is expected to work particularly well \cite{kaban2015new}.  
%\newpage  
%\nocite{langley00}

\bibliographystyle{unsrt}
\bibliography{paper-adaptive-arxiv}
\end{document}